\documentclass[11pt]{article}

\usepackage[]{emnlp2021}

\usepackage{times}
\usepackage{latexsym}

\usepackage[T1]{fontenc}
\usepackage[utf8]{inputenc}

\usepackage{microtype}
\usepackage{times}
\usepackage{helvet}
\usepackage{amsmath}
\usepackage{courier}
\usepackage{graphicx}
\usepackage{mathrsfs}
\usepackage{amsfonts}
\usepackage{CJKutf8}
\usepackage{ulem}
\usepackage{algorithm}
\usepackage{algorithmic}
\usepackage{float}
\usepackage{multirow}
\usepackage{multicol}
\usepackage{xcolor}
\usepackage{tabularx}
\usepackage{booktabs}
\usepackage{xspace}
\usepackage{subfigure}
\title{An Unsupervised Method for Building Sentence Simplification Corpora in Multiple Languages}

\newcommand{\ccnet}{\textsc{CCNet}\xspace}

\newcommand{\asset}{\textsc{ASSET}\xspace}
\newcommand{\turkcorpus}{\textsc{TurkCorpus}\xspace}

\newcommand{\easse}{\textsc{EASSE}\xspace}
\newcommand{\paranmt}{\textsc{ParaNMT-50M}\xspace}
\newcommand{\parabank}{\textsc{ParaBank}\xspace}
\newcommand{\alector}{\textsc{ALECTOR}\xspace}
\newcommand{\simplext}{\textsc{SIMPLEXT}\xspace}
\newcommand{\sacrebleu}{\textsc{SacreBLEU}\xspace}

\author{
    Xinyu Lu \thanks{~~Equal contribution.} ~~Jipeng Qiang\footnotemark[1] \thanks{~~Corresponding author.} ~~Yun Li ~~Yunhao Yuan ~~Yi Zhu\\
    Department of Computer Science, Yangzhou University, Jiangsu, China\\
    \texttt{\{181303216, jpqiang, liyun, yhyuan, zhuyi\}@yzu.edu.cn}
}

\begin{document}
\maketitle
\begin{abstract}

The availability of parallel sentence simplification (SS) is scarce for neural SS modelings. We propose an unsupervised method to build SS corpora from large-scale bilingual translation corpora, alleviating the need for SS supervised corpora. Our method is motivated by the following two findings: neural machine translation model usually tends to generate more high-frequency tokens and the difference of text complexity levels exists between the source and target language of a translation corpus. By taking the pair of the source sentences of translation corpus and the translations of their references in a bridge language, we can construct large-scale pseudo parallel SS data. Then, we keep these sentence pairs with a higher complexity difference as SS sentence pairs. The building SS corpora with an unsupervised approach can satisfy the expectations that the aligned sentences preserve the same meanings and have difference in text complexity levels. Experimental results show that SS methods trained by our corpora achieve the state-of-the-art results and significantly outperform the results on English benchmark WikiLarge.

\end{abstract}

\section{Introduction}

The task of sentence simplification (SS)  is to rephrase a sentence into a form that is easier to read and understand while conveying the same meaning \cite{chandrasekar1996motivations}. SS is first used as a preprocessing task of machine translation, and then is used to increase accessibility for those with cognitive disabilities such as aphasia \cite{carroll1998practical}, dyslexia \cite{rello2013simplify}, and autism \cite{evans2014evaluation}.  

Most popular methods \cite{wubben2012sentence,xu2016optimizing,zhang-lapata-2017-sentence,nisioi2017exploring,martin-etal-2020-controllable} have addressed SS as a monolingual machine translation task that translating from complex sentences to simplified sentences, whose performance rely heavily on the quality of parallel SS corpus. However, much work \cite{woodsend2011learning,coster2011learning,xu2016optimizing,qiang2019unsupervised} pointed out that the public English SS benchmark (WikiLarge \cite{zhang-lapata-2017-sentence}) which align sentences from English Wikipedia and Simple English Wikipedia are deficient, because they contain a large proportion of inaccurate or inadequate simplifications, which lead to SS methods that generalize poorly. Additionally, parallel SS corpus is difficult to obtain in all languages other than English. Therefore, in the paper, we focus on how to build SS corpora in multiple languages using an unsupervised method. 
\begin{figure}
    \centering
    \resizebox{\columnwidth}{!}{
    \includegraphics[width=85mm]{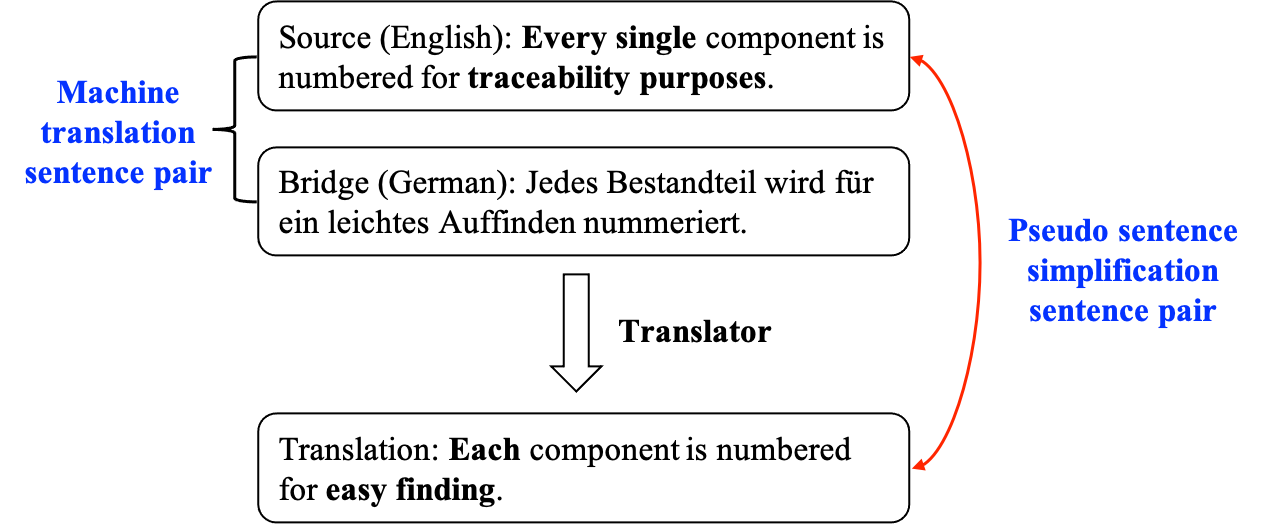}
    }
    \caption{Example of English sentence simplification pair generated by machine translation pair. Large-scale machine translation pairs (e.g. English-German) are chosen as a source base. The sentence of bridge language (e.g. German) is translated into an English sentence. After pairing the results of the source (English) and the translated sentence, we can harvest large-scale pseudo sentence pairs, as the red dashed arrow shows. }
    \label{fig1}
\end{figure}

Some work \citet{kajiwara2018text, martin2020multilingual} built pseudo parallel corpora by searching the nearest neighbor sentence for each sentence based on embedding model from a large-scale text corpus. We can see that the built corpora are more like paraphrase corpora instead of SS corpora. It is because: (1) It only guarantees that the aligned sentences are highly similar, and cannot guarantee that the aligned sentences preserve the same meanings; and (2) Each sentence pair does not distinguish between the simple sentence and the complex sentence. 

In this paper, we present an unsupervised method to build SS parallel corpora based on a large-scale bilingual translation corpus. Many languages have large-scale training corpora, which lie at the core of the recent success of neural machine translation (NMT) models. If we plan to build a pseudo English SS corpus, the main idea of our method is demonstrated in Figure \ref{fig1}: we use a translator to translate the sentence in a bridge language (e.g., German) into English, and pair them as a pseudo SS sentence pair. The idea is motivated by the following two findings \label{sec:findings}:

(1) NMT models usually tend to generate more high-frequency tokens and less low-frequency tokens \cite{gu-etal-2020-token, jiang2019improving}. Considering that the higher the word frequency, the more simple the word is, this phenomenon could be beneficial to text simplification \cite{saggion2017automatic,qiang2020AAAI}.

(2) The difference of text complexity levels exists between the source and target language of translation corpus \cite{bentz2016comparison}. From the perspective of linguistics, word entropy, morphological complexity, and syntactic complexity vary between languages. A sentence with lower complexity is more likely to be translated into a simpler one.

Each sentence pair in the pseudo SS corpus preserves the same meaning. We measure the difference of sentence complexity between the original sentence and the translated sentence using Flesch reading-ease score \cite{kincaid1975derivation}, and keep pairs with a higher complexity difference as SS corpus. For each remaining sentence pair, the sentence with a higher score will be treated as the simple sentence and the other sentence as the complex. 

The contributions of our paper are as follows:

(1) We propose an unsupervised method to build SS corpora in multiple languages because our method can be used to languages with large-scale NMT resources. Our method can guarantee that the aligned sentences preserve the same meanings and have difference in text complexity levels. 

(2) We provide SS corpora in three languages (English, French, and Spanish) to train SS models, alleviating the need for language-specific supervised corpora. We plan on making these resources publicly available after this paper is published\footnote{https://github.com/luxinyu1/Trans-SS}. 

(3) Experimental results show that SS methods on our English corpus significantly outperform the results on the English SS benchmark (WikiLarge). We adopt pre-trained language modeling BART on our English SS corpus to achieve the state-of-the-art in English with 42.69 SARI on \asset and 41.97 SARI on \turkcorpus datasets. 

\section{Related Work}

\subsection{Supervised Sentence Simplification}

Supervised sentence simplification (SS) methods treat sentence simplification task as monolingual machine translation task that translating from complex sentences to simplified sentences, requiring supervised parallel training corpora of complex-simple aligned sentences \cite{wubben2012sentence,martin-etal-2020-controllable,nisioi2017exploring,xu2016optimizing,zhang-lapata-2017-sentence,scarton2018learning,dong-etal-2019-editnts,qiang2020STTP}. The above methods have relied on WikiSmall \cite{zhu2010monolingual} or WikiLarge \cite{zhang-lapata-2017-sentence}, which aligned sentences from English Wikipedia and Simple English Wikipedia. The two datasets have been criticized \cite{woodsend2011learning,coster2011learning,xu2016optimizing,qiang2019unsupervised} because they contain a large proportion of inaccurate simplification (not aligned or only partially aligned) and inadequate simplification (not much simpler than complex sentence). Professional simplifications such as Newsela dataset \cite{xu-etal-2015-problems} have high-quality sentence pairs. But, it is usually accompanied by restrictive licenses that prevent widespread usage and reproducibility.

Researchers have attempted to design SS methods in other languages such as Spanish \cite{saggion2015making}, Portuguese \cite{aluisio2008towards}, Japanese \cite{goto2015japanese}, French \cite{gala2020alector} and Italian \cite{brunato2015design}. But, these approaches are limited by the availability of parallel SS corpora. In this paper, we propose a general framework that can be used to obtain large-scale SS data for these languages to train neural SS methods. 

\subsection{Unsupervised Sentence Simplification}

To overcome the scarcity of parallel SS corpus, unsupervised SS methods without using any parallel corpus have attracted much attention. Existing unsupervised SS methods can be divided into two classifications. The first scheme focuses on how to design an unsupervised SS method, and the second scheme concentrates on how to build a parallel SS corpus. 

\citet{2015Unsupervised} and \citet{kumar-etal-2020-iterative} are the pipeline-based unsupervised framework, where the pipeline of Narayan and Gardent is composed of lexical simplification, sentence splitting, and phrase deletion, the pipeline of Kumar et al. includes deletion, reordering, and lexical simplification. \citet{surya-etal-2019-unsupervised} proposed an unsupervised neural text simplification based on a shared encoder and two decoders, which only learn the neural network parameters from simple sentences set and complex sentences set. In other languages, there are unsupervised statistical machine translations for Japanese \cite{katsuta2019improving} and back-translation in Spanish and Italian \cite{palmero2019neural}. The performance of the above unsupervised SS methods is however often below their supervised counterparts.

Some work \cite{kajiwara2018text,martin2020multilingual} constructed SS corpora by searching the most similar sentences using sentence embedding modeling, and train SS methods using the constructed SS corpora. \citet{kajiwara2018text} calculated the similarity between the sentences from English Wikipedia by Word Mover's distance \cite{kusner2015word}. \citet{martin2020multilingual} adopted multilingual sentence embedding modeling LASER \cite{artetxe-etal-2018-unsupervised} to calculate the similarity between the sentences from 1 billion sentences from \ccnet \cite{Wenzek2019}. Since the aim of the two works is to find the most similar sentences from a large corpus, they cannot guarantee that the aligned sentences preserve the same meanings. 

\subsection{ Paraphrase Mining }

Some work has focused on generating paraphrase corpus for neural machine translation (NMT) systems using back-translation, where back-translation \cite{sennrich2015improving} is a technique widely used in NMT to enhance the target monolingual data during the training process. Specifically, the back-translation technique is used by translating the non-English side of bitexts back to English\cite{wieting2017learning} and pairing translations with the references. Two large paraphrase corpora (\paranmt\cite{wieting2017paranmt} and \parabank\cite{hu2019parabank}) are built based on this idea, and has been proven to have great potential in different translation-core tasks. Round-trip translation is also used in mining paraphrases \cite{mallinson2017paraphrasing} by translating sentences into another language then translating the result back into the original language. Similar to machine translation, back-translation is used to improve the performance of neural SS methods \cite{katsuta2019improving,palmero2019neural,qiang2019unsupervised}. \citet{mehta2020simplifythentranslate} trained a paraphrasing model by generating a paraphrase corpus using back-translation, which is used to preprocess source sentences of the low-resource language pairs before feeding into the NMT system. 

\begin{figure*}
    \resizebox{\textwidth}{!}{
    \centering
    \includegraphics{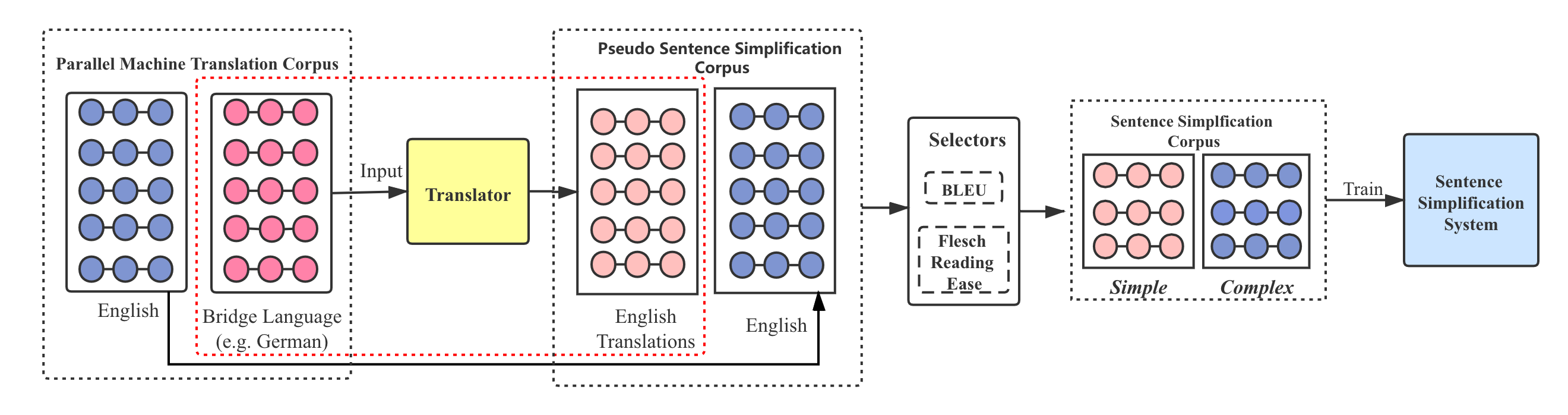}
    }
    \caption{The overview of our approach for Building English SS corpus. Our approach is composed of a high-resource bilingual translation corpus and a translator. A pseudo paraphrase corpus is synthesized by pairing the source sentences (English) and the translated sentences of the bridge language. Then, we select these complex-simple sentence pairs with a higher complexity difference, as the sentence simplification corpus. }
    \label{fig:approach}
\end{figure*}

The above work for building a large paraphrase corpus is to serve for NMT and other tasks, which is not fit for SS task. The difference of sentence complexity between the original sentence and the translated sentence for each sentence pair has not been taken into consideration, which is vitally important for SS task. Therefore, we focus on how to build a sentence simplification corpus, instead of a paraphrase corpus.

\section{Method}

In this paper, we present our unsupervised method to build SS corpora in multiple languages, which is motivated by high-frequency tokens generated by NMT modeling and the difference of text complexity levels between bilingual translation corpora. The overall architecture of the proposed method for building the English SS corpus is illustrated in Figure \ref{fig:approach}. Our method consists of two steps (Paraphrase Generation and Selectors) to build SS corpora, to achieve the following two requirements:

(1) The two sentences of each sentence pair should convey the same meaning. Given high-resource parallel machine translation corpus, we can obtain paraphrase corpus by translating the sentences of the bridge language into the target language using a Translator. 

(2) The two sentences of each sentence pair should have difference in text complexity levels. After obtaining the paraphrase corpus, we calculate the text complexity of the two sentences using text readability formulas and keep these pairs with a higher complexity difference. 

\paragraph{Pseudo SS Generation} In contrast to previous work\cite{wieting2017paranmt, hu2019parabank} mining paraphrases for focusing on lexical and sentence structure diversity, we mine paraphrases as pseudo SS corpus for mainly considering the fluency and syntax correctness. 

Specifically, if we plan to construct a SS corpus for a specific language $A$, we need to obtain a high-resource bilingual translation corpus of language $A$ and language $B$, and a Translator which can translate the sentences in language $B$ into the language $A$, where $B$ is the bridge language. It should be noted that the chosen bridge language determines the availability of parallel machine translation corpus, and also further determines the performance of the Translator. Therefore, the bridge language $B$ and the specific language $A$ should have a high-resource parallel corpus. For example, as shown in Figure \ref{fig:approach}, we choose German as a bridge language for building the English SS corpus.

\paragraph{Selectors} In this step, we design a simple pipeline consisting of only two selectors (BLEU and Flesch Reading Ease) for selecting some sentence pairs with a higher complexity difference.  

Firstly, we select these pairs with the BLEU scores\footnote{The BLEU scores in this work are calculated by \sacrebleu\cite{post2018call}} above a threshold $h_{\textit{BLEU}}$, for ensuring the quality of paraphrases. It is worthy to note that, this selector is mainly used to filter the unaligned sentence pairs in paraphrase corpus because translation mistakes are relatively rare owing to recent improvements on NMT models. Meanwhile, we filter out the translations which are the same as references.

Secondly, we measure the difference of text complexity using Flesch reading ease score (FRES) \cite{kincaid1975derivation}, which is designed to indicate how difficult a sentence is to understand, and is widely used to evaluate the performance of SS. FRES grades the text from 0 to 100. The higher scores indicate the sentences are easier to read. As usual, the difference of one school grade level in FRES is 10, e.g., 5th grade (100.00-90.00) and 6th grade (90.0-80.0). The formula of FRES is,

\begin{equation}
\resizebox{.89\hsize}{!}{
$
    k_{1}-k_{2}\left(\frac{\text {\# words }}{\text {\# sentences }}\right)-k_{3}\left(\frac{\text {\# syllables }}{\text {\# words }}\right)
$
}
\label{sec:fres}
\end{equation}

Here $k_{1}$, $k_{2}$, $k_{3}$ are coefficients that vary in different languages. 

To ensure simplicity, we only keep the sentence pairs with a FRES difference higher than a threshold $h_{\textit{FRES}}$. In our experiments, we set $h_{\textit{BLEU}}$ = 15.0 and $h_{\textit{FRES}}$ = 10.0, where $h_{\textit{FKE}}=10.0$ means that for each sentence pair, the simplified version should be at least one school level simpler than its its unsimplified counterpart.

\section{Sentence Simplification Corpora}

\begin{table*}[h]
\resizebox{\textwidth}{!}{
\centering\small
\begin{tabular}{l|l}\toprule
\textbf{Bridge Language} & Er sagt er bekomme Platzangst und fuehle sich, als ob er in einem Sarg begraben werde. \\
\textbf{Complex} & He says he gets \textbf{claustrophobic}, that he feels \textbf{trapped as if} he was buried in a coffin.\\
\textbf{Simple} & He says he gets \textbf{scared} and feels \textbf{like} he's being buried in a coffin.\\
\midrule
\textbf{Bridge Language} & Das Hotel Gates am Kudamm, mit seiner einmaligen Gastfreundschaft, müssen Sie unbedingt einmal selbst erleben. \\
\textbf{Complex} & \textbf{You simply must experience} the Hotel Gates Am Kudamm with its unique \textbf{concept} of hospitality. \\
\textbf{Simple} & The Hotel Gates Am Kudamm, with its unique hospitality, is a \textbf{must-see}. \\
\midrule
\textbf{Bridge Language} & Das Geld muss in Unternehmen investiert werden, die garantieren, dass Hochschulabgänger einen Arbeitsplatz finden. \\
\textbf{Complex} & The money must be invested in \textbf{enterprises} which guarantee that graduates will find \textbf{employment}.\\
\textbf{Simple} & The money must be invested in \textbf{companies} that guarantee that graduates will find \textbf{a job}.\\\bottomrule
% \textbf{Bridge Language} & Jedes Bestandteil wird für ein leichtes Auffinden nummeriert. \\
% \textbf{Complex} & \textbf{Every single} component is numbered for \textbf{traceability purposes}. \\
% \textbf{Simple} & \textbf{Each} component is numbered for \textbf{easy finding}. \\
% \hline
% \textbf{Bridge Language} &  \\
% \textbf{Complex} & Whether we like it or not, nuclear energy provides a solution for both problems. \\
% \textbf{Simple} & Nuclear energy, like it or not, solves both problems. \\
% \hline
\end{tabular}
}
\caption{Examples of English SS corpus generated by our method. The differences between the complex sentence and the simple sentence are emphasized in bold .}
\label{examples}
\end{table*}

\begin{table}[h]
\centering\small
\resizebox{\columnwidth}{!}{
\begin{tabular}{l|c|ccc} \toprule
& \textbf{WikiLarge} & \textbf{English} & \textbf{French} & \textbf{Spanish} \\\midrule
\textbf{Vocab(complex)} & 169,349 & 196,301 & 112,335 & 119,876\\
\textbf{Vocab(simple)} & 135,607 & 165,130 & 102,672 & 104,361\\ \midrule
\textbf{Avg(complex)} & 21.93 & 18.95 & 26.17 & 28.00\\
 \textbf{Avg(simple)} & 16.14 & 19.36 & 27.74 & 25.79\\ \midrule
 \textbf{Total pairs} & 296,402 & 816,058 & 621,937 & 487,862\\ \bottomrule
\end{tabular}
}
\caption{\label{tab:profile} Statistics of our building corpora in English, French, and Spanish compared with Wikilarge. Avg(complex) and Avg(simple) are the average numbers of words in the complex sentences and the simpler sentences, respectively. }
\end{table}

Our unsupervised method can be used to languages with large-scale bilingual translation corpora. According to this principle, we choose the three languages (English, French, and Spanish) to build SS corpora, to train SS systems.

\paragraph{English} In the step of paraphrase generation, we choose German as bridge language and obtains the bilingual translation corpus provided by huggingface\footnote{https://cdn-datasets.huggingface.co/translation/wmt\_en\_\\de.tgz} with 4,000,000 sentence pairs as \textit{De-En}. We use Facebook FAIR’s state-of-the-art \textit{De $\to$ En} model\footnote{https://dl.fbaipublicfiles.com/fairseq/models/wmt19.en-de.joined-dict.ensemble.tar.gz}\cite{ng2019facebook} as the translator, which is based on big Transformer\cite{vaswani2017attention} architecture training on WMT19 dataset. The parameters of FRES in English are set to $k_{1}=206.835, k_{2}=1.015, k_{3}=84.6$. 

\paragraph{French and Spanish} For both French and Spanish, English is chosen as the bridge language. The bilingual translation corpora for the two languages are from the full Europarl-v7 dataset\footnote{https://www.statmt.org/europarl/}, where the sentence pairs of English-French and English-Spanish are 1,965,734 and 2,007,723, respectively.

For French, the translator from English to French is also a Transformer-based model\footnote{https://dl.fbaipublicfiles.com/fairseq/models/wmt14.en-fr.joined-dict.transformer.tar.bz2}\cite{ott-etal-2018-scaling}. For Spanish, the translator from English to Spanish is fine-tuned by pre-trained language model mBART\cite{liu2020multilingual}  \footnote{https://dl.fbaipublicfiles.com/fairseq/models/mbart/mbart\\.cc25.v2.tar.gz\label{fn:mbart}}. 

 The parameters of FRES in French is set to $k_{1}=207, k_{2}=1.015, k_{3}=73.6$, and the parameters in German are $k_{1}=180, k_{2}=58.5, k_{3}=1.0$.

\paragraph{Statistics and Examples} We show some examples of the sentence pairs generated by our method in Table \ref{examples}. We report the statistics of our building corpora in Table \ref{tab:profile}. The numbers of sentence pairs in English, French, and Spanish are 816,058, 621,937, and 48,7862, respectively. Because the SS task is a paraphrase generation task using easier words, the length of the complex sentence and the simple sentence are roughly the same, and the size of the vocabulary in the simple sentence set should be smaller than the complex sentence set. From Table \ref{tab:profile}, we can see that our three corpora satisfy the expectations of the SS task. In contrast to our corpora, the length of the complex sentence in WikiLarge is longer than the simple sentence, because it focuses on the deletion of content.

\section{Experiments}

We design experiments to answer the following two questions:

\noindent\textbf{Q1. Effectiveness:} Is the English SS corpus built by our method a better dataset compared with the benchmark WikiLarge and the dataset built by \cite{martin2020multilingual}?

\noindent\textbf{Q2. Universality:} Can our unsupervised method be used to build SS corpora for other languages with large-scale bilingual translation corpora?

\begin{table*}[h]
\centering\small
\resizebox{\textwidth}{!}{
\begin{tabular}{llcccccc}\\\toprule
 & \textbf{Data} & \multicolumn{3}{c}{\textbf{\turkcorpus}} & \multicolumn{3}{c}{\textbf{\asset}} \\
 & & SARI $\uparrow$ & FKGL $\downarrow$ & BLEU $\uparrow$ & SARI $\uparrow$ & FKGL $\downarrow$ & BLEU $\uparrow$ \\\midrule
Source & --- & 26.29 & 10.02 & 99.36 & 20.73 & 10.02 & 92.81  \\
Reference  & --- & 40.21 & 8.73 & 73.00 & 45.14 & 6.48 & 70.12  \\\midrule
PBMT-R\cite{wubben-etal-2012-sentence} & WikiSmall & 38.04 & 8.85 & 82.49 & 34.63 & 8.85 & 79.39  \\
Dress-LS\cite{zhang-lapata-2017-sentence} & WikiLarge & 36.97 & 7.66 & 81.08 & 36.59 & 7.66 & 86.39  \\
DMASS-DCSS\cite{zhao-etal-2018-integrating} & WikiLarge & 39.92 & 7.73 & 73.29 & 38.67 & 7.73 & 71.44  \\
ACCESS\cite{martin-etal-2020-controllable} & WikiLarge & 41.38 & 7.29 & 76.36 & 40.13 & 7.29 & 75.99 \\\midrule
UNTS\cite{surya-etal-2019-unsupervised} & Unsupervised & 36.29 & 7.60 & 76.44 & 35.19 & 7.60 & 76.14 \\
BTTS10\cite{kumar-etal-2020-iterative} & Unsupervised & 36.91 & 7.83 & 82.00 & 35.72 & 7.83 & 83.01 \\\midrule
LSTM & WikiLarge & 35.69 & 7.7 & 79.45 & 35.81 & 6.06 & 72.3 \\
& \textbf{Ours} & 38.21 & 8.41 & 76.85 & 37.65 & 7.97 & 71.71 \\\midrule
ConvS2S & WikiLarge & 36.83 & 7.58 & 80.40 & 36.48 & 7.18 & 82.29 \\
& \textbf{Ours}  & \textbf{38.98} & 8.66 & 73.79 & \textbf{37.92} & 7.89 & 69.67 \\\midrule
Transformer & Wikilarge & 37.05 & 8.42 & 86.71 & 34.16 & 8.42 & 84.47 \\
& MUSS\cite{martin2020multilingual} & 38.06 & 9.43 & 63.70 & 38.03 & 9.41 & 61.76 \\
& \textbf{Ours} & 39.99 & 7.97 & 72.75 & 39.58 & 7.83 & 70.81 \\\midrule
% BART$_{large}$+ACCESS & \textbf{P.}\cite{martin2020multilingual} & 63.76 & 40.85 & 8.79 & 66.23 & 42.65 & 8.23 \\
% BART$_{large}$+ACCESS & \textbf{S.}\cite{martin2020multilingual} & --- & --- & --- & 60.18 & 42.44 & 7.18 \\
BART & WikiLarge & 38.96 & 8.15 & 84.58 & 36.81 & 8.15 & 85.66 \\
 & MUSS\cite{martin2020multilingual} & --- & --- & --- & 39.73 & 9.26 & 65.00 \\
& \textbf{Ours} & \textbf{41.97} & 8.21 & 73.72 & \textbf{42.69} & 7.94 & 71.83 \\
\bottomrule
\end{tabular}
}
\caption{\label{table:english_results} Results of English sentence simplification. $\uparrow$The higher, the better. $\downarrow$The lower, the better. $-$ indicates the results that are not found in the original paper.}
\end{table*}

\subsection{Evaluation Datasets}

We choose four datasets to evaluate the performance of SS modelings on our corpora. The statistics are reported in Table \ref{tab:valid}.

\begin{table}[h]

\centering\small

\begin{tabular}{lcccc}
\toprule
& \textbf{Lang.} & \textbf{\#Valid} & \textbf{\#Test} & \textbf{C.R.}  \\ \midrule
\turkcorpus & English & 2000 & 359 & 0.95 \\
\asset & English& 2000 & 359 & 0.83  \\
\alector & French & 800 & 801 & 0.97  \\
\simplext & Spanish & 708 & 708 & 0.48  \\ \bottomrule
\end{tabular}
\caption{\label{tab:valid} The statistics of SS evaluation datasets. \textbf{C.R.}(\textbf{C}ompression \textbf{R}atio) is the amount of compression of the complex sentence relative to the simple sentence. }

\end{table}

For evaluating English simplification task, we use two widely used evaluation benchmarks \turkcorpus \cite{xu2016optimizing} and \asset \cite{alva-manchego-etal-2020-asset} of WikiLarge dataset. Both \turkcorpus and its improved version \asset consist of 2,000 valid sentences and 359 test sentences. Each original sentence in \turkcorpus has 8 simplification references collected through Amazon Mechanical Turk.  \asset with 10 simplification references per original sentence focuses on multiple simplification operations including lexical paraphrasing, compression, and sentence splitting. 

For French, we use \alector\cite{gala2020alector} for evaluation, which contains 1601 sentence pairs. It contains 79 original literary and scientific texts along with their simplified equivalents, which are chosen from materials for French Primary school students. We split it into a valid set (first 800 pairs) and a test set (next 801 pairs).
 
For Spanish, we use \simplext\cite{saggion2015making,saggion2017automatic} for evaluation, which contains 1416 sentence pairs. It is from 200 news articles that were manually simplified by experienced experts for people with learning disabilities. We split it into a valid set (708 pairs) and a test set (708 pairs).

\subsection{Training Details}

To compare the quality of our building corpora with other training datasets, we test the following four models: LSTM-based, ConvS2S-based, Transformer-based, and BART-based models. We implement the four models via fairseq\cite{ott-etal-2019-fairseq}. The parameters of all these models are tuned with SARI on validation sets. 

For BART-based model used for English SS, we initialize the model with the pretrained weights\footnote{https://dl.fbaipublicfiles.com/fairseq/models/bart.large.ta\\r.gz}. For BART-based model used for French and Spanish languages, we adopt a multilingual pretrained BART (mBART) with the weights\textsuperscript{\ref{fn:mbart}} pretrained on 25 languages.

We adopt the Adam optimizer with $ \beta_{1}=0.9 $, $ \beta_{2}=0.999 $, $ \epsilon = 10^{-8} $ for LSTM-based, Transformer-based and BART-based models, the NAG optimizer for ConvS2S-based model. Dropout is set 0.1 for LSTM-based, ConvS2S-based and BART-based models and 0.2 for Transformer-based model. The initial learning rate are set to $5 \times 10^{-4}$, $3 \times 10^{-4}$, $lr=3 \times 10^{-5}$ for  LSTM-based, Transformer-based and BART(large )-based models, respectively. We use a fixed learning rate of $lr=0.5$ for ConvS2S-based model. Byte Pair Encoding(BPE) is used in all the models for word segmentation.

\subsection{Evaluation Metrics}

SARI\cite{xu2016optimizing} is the main metric to evaluate text simplification models, which calculates the arithmetic mean of the $n$-gram F1 scores of three operations (keeping, adding, and deleting) through comparing the generated sentences to multiple simplification references and the original sentences.

Flesch-Kincaid Grade Level (FKGL)\cite{kincaid1975derivation} based on FRES (Formula \ref{sec:fres}) is widely used to evaluate the SS task, which measures the readability of the system output.

Earlier work also used BLEU \cite{papineni2002bleu} as a metric, but recent work has found that it does not correlate with simplicity \cite{sulem-etal-2018-bleu}. Systems with high BLEU scores are thus biased towards copying the original sentences as a whole (e.g., 99.36 on \turkcorpus or 92.81 on \asset). For completeness, we also report BLEU scores. For the above metrics, we use standard simplification evaluation tool \easse \footnote{https://github.com/feralvam/easse} to compute their scores.

\subsection{English Simplification}

We choose four supervised SS methods (PBMT-R\cite{wubben-etal-2012-sentence}, Dress-LS\cite{zhang-lapata-2017-sentence}, DMASS-DCSS\cite{zhao-etal-2018-integrating}, and ACCESS\cite{martin-etal-2020-controllable} ), two unsupervised SS methods (UNTS\cite{surya-etal-2019-unsupervised} and BTTS10\cite{kumar-etal-2020-iterative}) to compare. We also choose the SS corpus built by \cite{martin2020multilingual} as a comparison. 

Table \ref{table:english_results} summarizes the evaluation results of SS methods on our building English corpus. We first compare the results between our building English SS corpus and English SS corpus WikiLarge. In terms of SARI metric, we can see that the four models (LSTM-based, ConvS2S, Transformer-based, and BART-based) on our data significantly outperform the results on WikiLarge dataset, demonstrating the promise of building SS corpora using our method. BART-based method achieves the best results compared with the other three models (LSTM-based, ConvS2S, and Transformer-based). On \turkcorpus and \asset, BART-based method on our data significantly outperforms the results on WikiLarge by a large margin (+3.01, +5.88 SARI). In terms of readability, BART on our data obtains lower (=better) FKGL compared to the results on WikiLarge. We believe this improvement shows that our method for building English SS corpus is a good choice for SS task. 

We then compare the results between our building corpus and MUSS build by \cite{martin2020multilingual}. Compare with MUSS, SS methods on our dataset outperform the results on MUSS in terms of all the measurements. Transformer-based method on our dataset achieves a large improvement of (+1.93, + 1.55 SARI) on \turkcorpus and \asset, and BART-based method achieves a large margin of +2.96 SARI on \asset. In terms of readability, Transformer-based and BART-based methods obtain lower FKGL compared with MUSS, which indicated the output of the SS methods is easier to understand. These indicate that the effectiveness of our method on building SS English corpus. 

\subsection{French and Spanish Simplification}

\begin{table}[h]
\centering\small
\resizebox{\columnwidth}{!}{
\begin{tabular}{llcccc}
\toprule
& \textbf{Data} & \multicolumn{2}{c}{\textbf{\alector}} & \multicolumn{2}{c}{\textbf{\simplext}} \\
&  & SARI $\uparrow$ & FRES $\uparrow$  & SARI $\uparrow$ & FRES $\uparrow$ \\ \midrule
Source & --- & 26.36 & 66.57  & 5.65 & 49.40  \\
Pivot & --- & 38.52 & \textbf{65.55}  & 25.98 & 56.30   \\\midrule
mBART+ & MUSS & 38.35 & 68.36  & 19.81 & 55.07\\
\midrule
Transformer & \textbf{Ours} & 36.93 & 75.56  & \textbf{30.37} & 44.51 \\
mBART & \textbf{Ours} & \textbf{39.00} &  73.15  & 27.83 & \textbf{47.30}\\
\bottomrule
\end{tabular}
}

\caption{\label{table:french_and_spanish} Results of unsupervised sentence simplification in French and Spanish. We choose FRES metric instead of its revision FKGL in French and Spanish because the coefficients of FKGL in these two languages are not available. }
\end{table}

Our approach can be applied to any language owing to large-scale translation corpora. Different from English SS task, large-scale SS training corpus in other languages is hard to obtain. For a better comparison, we add one new baseline (Pivot) via machine translation. Specifically, for Pivot, give one non-English sentence, we translate the sentence to English and translate the translated sentence back into the source language. Here, we use Google Translator \footnote{https://translate.google.com/} for French and Spanish translation. We also choose the best results of mBART+ (mBART+ACCESS) on MUSS dataset \cite{martin2020multilingual}, where ACCESS \cite{martin-etal-2020-controllable} is a control mechanism to the parameters of SS model by controlling attributes such as length, lexical complexity, and syntactic complexity. 

The results are shown in Table \ref{table:french_and_spanish}. We can see that the same SS methods on our dataset outperform the results on MUSS, which verifies that our method is more fit for SS task. Compared with the results of Pivot, SS methods on our building SS corpora can generate more simplified sentences. We conclude that our method for building SS corpora can be used to languages with large-scale bilingual translation corpora.

\subsection{Ablation Study of our method}

To further analyze the factors affecting our unsupervised method for building SS corpus, we do more experiments in this subsection.

\begin{table}[h]
    \centering\small
    \resizebox{\columnwidth}{!}{
    \begin{tabular}{llcccc} \\\toprule
    & \textbf{Method} & \multicolumn{2}{c}{\textbf{\turkcorpus}} & \multicolumn{2}{c}{\textbf{\asset}} \\
    & & SARI $\uparrow$ & FKGL $\downarrow$ & SARI $\uparrow$ & FKGL $\downarrow$ \\
    \midrule
    Pseduo SS & Transformer & 34.18 & 9.49 & 29.46 & 9.49 \\
    & BART & 33.94 & 9.72 & 29.92 & 9.72 \\
    \midrule
    w/o BLEU & Transformer & 38.53 & 6.37 & 39.05 & 6.08 \\ 
                & BART & 38.61 & 7.01 & 40.62 & 6.54 \\
    w/o FRES  & Transformer & 34.86 & 9.88 & 30.49 & 9.63 \\
                & BART & 35.97 & 9.69 & 31.54 & 9.69 \\
    \midrule
    \textit{full} & Transformer & 39.99 & 7.97 & 39.58 & 7.83 \\
    & BART & 41.97 & 8.21 & 42.69 & 7.64 \\\bottomrule
    \end{tabular}
    }
\caption{\label{tab:ablations} Ablation study results of SS methods on our English corpus without BLEU selector and FRES selector. "w/o" denotes "without".}
\end{table}

\begin{table*}[h]
\centering\small
\begin{tabular}{ll}\\\toprule
Source & He was diagnosed with \textbf{inoperable abdominal} cancer in April 1999.\\
Reference & He was diagnosed with abdominal cancer in April 1999.\\
Ours & He was diagnosed with \textbf{stomach} cancer in April 1999.\\\midrule
Source & Heavy rain fell \textbf{across portions of Britain} on October 5, causing localized \textbf{accumulation of flood waters}.\\
Reference & Heavy rain fell across Britain on October 5, causing accumulation of flood waters. \\
Ours & Heavy rain fell \textbf{on parts of the UK} on October 5, causing localized \textbf{flooding}.\\\midrule
Source & Admission to Tsinghua is \textbf{extremely} competitive.\\
Reference & Admission to Tisinghua is competitive.\\
Ours & Admission to Tsinghua is \textbf{very} competitive. \\\midrule
Source & They are \textbf{culturally akin} to the coastal peoples of Papua New Guinea. \\
Reference & They are \textbf{similar to} the coastal peoples of Papua New Guinea. \\
Ours & They are \textbf{similar in culture} to the coastal peoples of Papua New Guinea. \\

\bottomrule
\end{tabular}
\caption{\label{tab:genexamples}Examples of simplifications generated by BART-based method on our building English corpus. The bold words highlight the differences.}
\end{table*}

\paragraph{(1) Influence of each selector in our method. }

To evaluate the effect of each selector in our methods, we build four different SS corpus: pseudo SS corpus, the corpus building by our method without BLEU selector, the corpus building by our method without FRES selector, and the corpus building by our full method. We choose two SS methods (Transformer-based and BART-based) to do the experiments, and the results show in Table \ref{tab:ablations}. It is very obvious that the results on pseudo SS corpus are the worst and our method combing two selectors achieves the best results. FRES selector in our method is more important than BLEU selector, because FRES selector is used to select the sentence pairs with a higher complexity difference and BLEU selector is only used to filter the unaligned sentence pairs.

\paragraph{(2) Influence of the size of the corpus built by our method. }

\begin{figure}
    \resizebox{\columnwidth}{!}{
    \centering
    \includegraphics{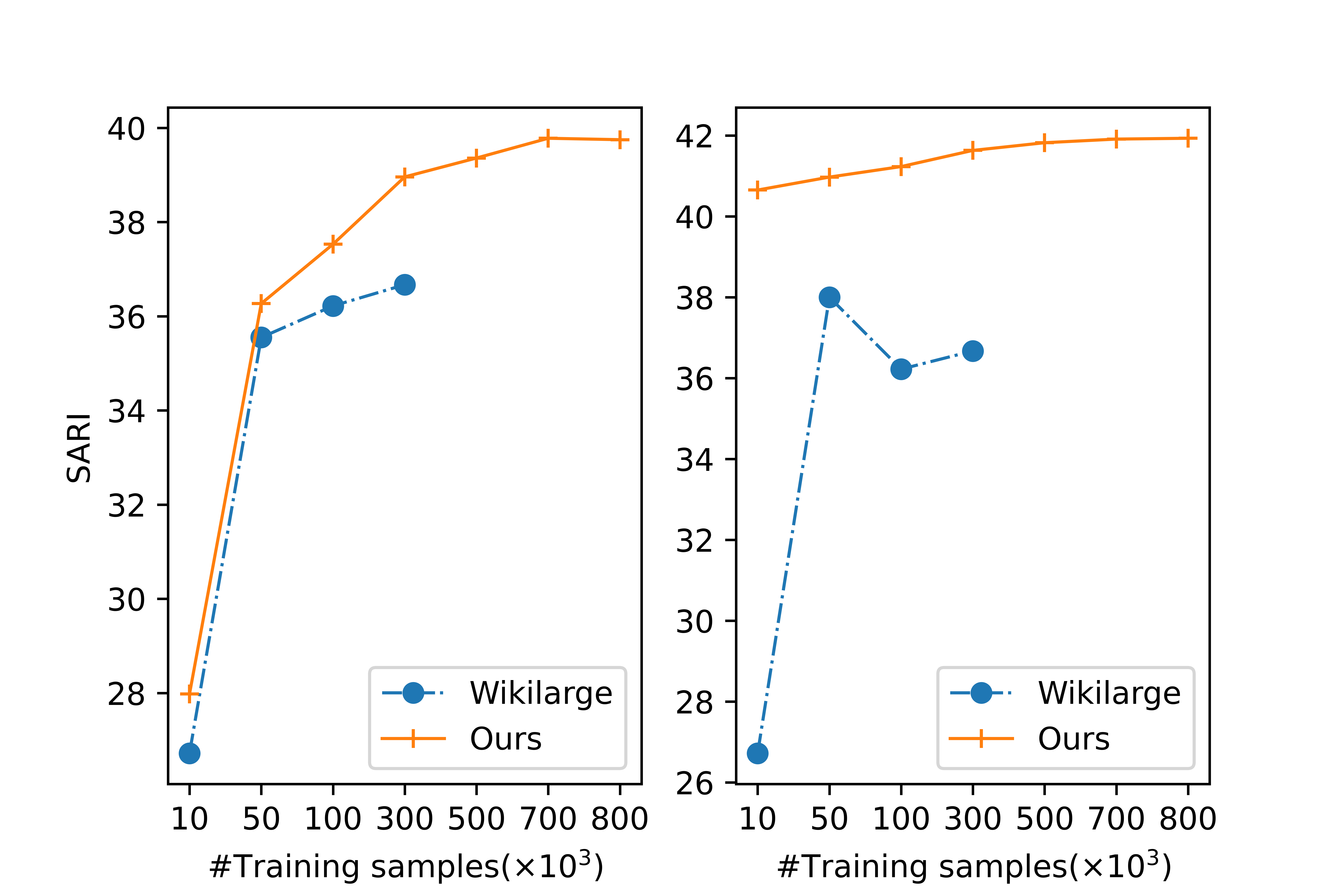}
    }
    \caption{\label{fig:numab}The performance of Transformer-based and BART-based method on \turkcorpus when varying the size of the corpus.}
\end{figure}

Because the size of our SS corpus is 816,058 and the size of WikiLarge is 296,402, we vary the size from 10K to 800K to analyze the results of Transformer-based on the two corpora.  Due to the size of WikiLarge, we only show the results of WikiLarge in the first 30K samples. We can see that the SARI values increase at first and keep stable finally when increasing the size of training samples. We see that the size of the SS corpus is of vital importance for SS methods. In the paper, we only choose a bilingual translation corpus of size 4,000,000. In the future, we will try to build SS corpora using a more large-scale bilingual translation corpus.

\subsection{Qualitative Study}

Table \ref{tab:genexamples} shows some simplified sentences from the test set of \turkcorpus by our method BART-based method trained with our building English corpus. Our model reduces more linguistic complexity of the source sentence, while still retaining its original information and meaning. We can found that our method more focuses on lexical simplification, e.g., "stomach" as a simpler for "inoperable abdominal", "very" as a simpler for "extremely", etc.  We draw the same conclusions from these examples that our building method can be used to train SS methods.

\section{Conclusions}

We propose an unsupervised method to build large parallel corpora for training sentence simplification (SS) models. Our method consists of a high-resource bilingual translation corpus and a translator. Unsupervised SS models can be trained by pairing the source sentences in the bilingual translation corpus and the translated sentences of the bridge language generated by the translator. We conduct experiments and show that SS models trained on synthetic data generated by our approach significantly outperform the results on English benchmark WikiLarge. In the future, we plan to investigate the influence of different text readability methods.

\bibliography{emnlp2021}
\bibliographystyle{acl_natbib}

\end{document}